\documentclass[10pt,letterpaper]{article}

\usepackage{hyperref}
\hypersetup{
    colorlinks=true,
    linkcolor=black,      
    citecolor=black,      
    filecolor=black,      
    urlcolor=blue         
}
\usepackage[table]{xcolor}
\usepackage{cogsci}
\usepackage{algorithmic}
\usepackage{algorithm}
\usepackage{graphicx}
\usepackage{colortbl}
\usepackage{comment}
\usepackage[most]{tcolorbox}
\usepackage{enumitem} 
\usepackage{amsmath}
\usepackage{subcaption} 
\usepackage{float}      
\usepackage{xspace}
\usepackage{caption}
\usepackage{csquotes}
\usepackage{placeins}
\usepackage{amsmath} 
\usepackage{amssymb} 
\usepackage{cleveref}
\usepackage{framed}
\usepackage{lipsum} 
\usepackage{multirow}
\usepackage{booktabs}
\usepackage{nameref}

\definecolor{LightCyan}{rgb}{0.318, 0.439, 0.843}
\definecolor{LightCyan}{rgb}{0.8295, 0.85975, 0.96075}
\definecolor{IncrGreen}{rgb}{0.25, 0.55, 0.15}
\definecolor{DecrRed}{rgb}{0.55, 0.25, 0.25}
\newcommand{\qinc}[1]{\normalsize\textcolor{IncrGreen}{$\uparrow #1$}}

\cogscifinalcopy 

\newcommand*{\gemini}{\textsc{gemini-1.5-flash}\xspace}
\newcommand*{\llama}{\textsc{Llama-3.2-8B}\xspace}
\newcommand*{\chatgpt}{\textsc{ChatGPT-4o}\xspace}
\usepackage{pslatex}
\usepackage{apacite}
\usepackage{float} 

\usepackage[symbol]{footmisc}
\usepackage{bigfoot}
\DeclareNewFootnote{default}
\MakeSortedPerPage{footnote}

\renewcommand{\thefootnote}{\arabic{footnote}}

\title{Complexity in Complexity: \\ Understanding Visual Complexity Through Structure, Color, and Surprise}
 
\author{{\large \bf Karahan Sarıtaş$^{1}$,  \large \bf  Peter Dayan$^{1,2}$, Tingke Shen$^{2}$$^\ast$, Surabhi S Nath$^{1,2,3}$$^\ast$} \\
  $^1$University of Tübingen, Tübingen, Germany\\
  $^2$Max Planck Institute for Biological Cybernetics, Tübingen, Germany\\
  $^3$Max Planck School of Cognition, Leipzig, Germany
  }

\begin{document}

\maketitle

\begin{abstract}
Understanding how humans perceive visual complexity is a key area of study in visual cognition. Previous approaches to modeling visual complexity assessments have often resulted in intricate, difficult-to-interpret algorithms that employ numerous features or sophisticated deep learning architectures. While these complex models achieve high performance on specific datasets, they often sacrifice interpretability, making it challenging to understand the factors driving human perception of complexity. Recently (Shen, et al. 2024) proposed an interpretable segmentation-based model that accurately predicted complexity across various datasets, supporting the idea that complexity can be explained simply. In this work, we investigate the failure of their model to capture structural, color and surprisal contributions to complexity. 
To this end, we propose Multi-Scale Sobel Gradient (MSG) which measures spatial intensity variations, Multi-Scale Unique Color (MUC) which quantifies colorfulness across multiple scales, and surprise scores generated using a Large Language Model.
We test our features on existing benchmarks and a novel dataset (Surprising Visual Genome) containing surprising images from Visual Genome. Our experiments demonstrate that modeling complexity accurately is not as simple as previously thought, requiring additional perceptual and semantic factors to address dataset biases. Our model improves predictive performance while maintaining interpretability, offering deeper insights into how visual complexity is perceived and assessed. Our code, analysis and data are available at our Github repository\footnote[1]{\url{https://github.com/Complexity-Project/Complexity-in-Complexity}}.

\end{abstract}

\renewcommand{\thefootnote}{$\ast$}
\footnotetext{Joint supervision}
\renewcommand{\thefootnote}{\arabic{footnote}}
\section{Introduction}

Visual complexity is a fundamental attribute of images that reflects the level of detail, intricacy, and variation in visual elements within a scene \cite{Snodgrass1980ASS}. This perceptual characteristic encompasses multiple dimensions, including the number of lines, density of elements \cite{mcdougall1999measuring}, quantity of objects \cite{Olivia2004IdentifyingTP}, clutter \cite{KYLEDAVIDSON2023105319}, symmetry \cite{KYLEDAVIDSON2025108525}, and variety of colors \cite{corchs2016predicting}. 
Understanding visual complexity is crucial across domains, from user interface design \cite{10.1145/3206505.3206549, AKCA2021102031} to cognitive psychology \cite{10.1007/978-3-642-02728-4_17, madan2018visual}. While deep learning models predict complexity well, their black-box nature \cite{li2022interpretabledeeplearninginterpretation} limits interpretability, which is essential for applications like education \cite{Stoesz, Ghai} and information visualization \cite{Zhu2007, 10.1007/978-3-540-76858-6_56}. Developing interpretable, perception-aligned features remains a key research goal.

A recent study by \cite{shen2024simplicitycomplexityexplaining} bridged the gap between handcrafted features and deep learning approaches by proposing a simple two-feature model based on outputs from deep segmentation models (the numbers of segments and semantic classes in the image). Their simple, yet interpretable, model achieved superior performance compared to baseline on many naturalistic and art datasets, supporting the idea that the numbers of segments and classes are generic features that explain complexity well. However, their model fails to take into account the structure and arrangement of components in the image. 

Here, we build upon the model from  \cite{shen2024simplicitycomplexityexplaining}, by studying two of its major failure modes. First, we find that datasets with high structural and color regularity require two low-level features to explain complexity: Multi-Scale Sobel Gradient (MSG) and Multi-Scale Unique Color (MUC). MSG captures continuous spatial intensity variations across multiple scales, offering a richer representation of image structure. MUC quantifies color diversity also at multiple scales, and at different color resolutions. The latter feature builds on the concept of \enquote{colorfulness} from \cite{50×70}, and provides a more robust and performant measure of chromatic complexity.


We discovered a second failure mode of \cite{shen2024simplicitycomplexityexplaining} which concerns the rather underexplored contribution of whole-image, holistic or emergent information to visual complexity. Previous research by \cite{Forsythe} demonstrated that participants in a visual complexity experiment exhibited a familiarity bias, perceiving familiar shapes as less complex compared to unfamiliar or novel ones. Similarly, \cite{Snodgrass1980} found a negative correlation between visual complexity and familiarity in their experiment. \cite{10.1007/978-3-642-02728-4_17} suggests that complexity judgments are context-dependent and that familiarity should be considered when developing a model of visual complexity. Motivated by their findings, we show that holistic, whole-image surprise judgments of scenes contribute significantly to the visual complexity of naturalistic images. We define surprising images as those containing unusual, or contextually novel elements, i.e., the opposite of familiarity. \cite{meyer1997surprise, Ortony}. To this end, we introduce a novel dataset, SVG, containing both surprising and randomly sampled images from the well-studied Visual Genome dataset \cite{krishna2016visualgenomeconnectinglanguage} and show that surprise scores generated by a large language model (LLM) explains significant variance in the perceived complexity of images in this dataset.

\section{Methods}

\subsection{Modeling and Evaluation}

In this work, we evaluate the effectiveness of sets of features to explain visual complexity on datasets using linear regression models. We follow the procedure in \cite{shen2024simplicitycomplexityexplaining}, and fit $M$ repetitions of 3-fold cross-validated linear regression. $M$ is determined by the size of the dataset to address statistical variability in smaller datasets. We measure performance using the mean Spearman correlation coefficient across all test splits. We compare the performance of our models to the same baselines in \cite{shen2024simplicitycomplexityexplaining}: handcrafted features \cite{corchs2016predicting, KYLEDAVIDSON2023105319} and deep learning approaches \cite{SARAEE2020102949, feng2023ic9600}. 

\subsection{Datasets}

We evaluate our method on four publicly available datasets with human-rated complexity scores: RSIVL (49 images) (Corchs et al., 2016), VISC (800 images) (Kyle-Davidson et al., 2023), Savoias (1400 images across 7 categories) (Saraee et al., 2020), and IC9600 (9600 images across 8 categories) (Feng et al., 2022). We select image subsets to demonstrate the effectiveness of our new features to explain complexity - VISC and IC9600 architecture subset for MSG, Savoias Art and Suprematism subsets for MUC, and Savoias Interior Design and IC9600 Abstract for the combination of MSG and MUC. Finally, we use our novel dataset SVG to show that surprisal contributes significantly to complexity.

\subsection{SVG: A Dataset of Surprising Images}

To the best of our knowledge, there is currently no existing image set with surprising images that can be used to study visual complexity. To fill this gap, we introduce a new dataset containing 100 highly surprising images and 100 (on average, less surprising) images, the latter of which were sampled uniformly at random from the Visual Genome, subject to stratification according to bins of complexity based on IC9600 scores \cite{feng2023ic9600}. To confirm that our subjectively selected images are more surprising than randomly sampled ones, we compare the distributions of LLM-generated surprise scores (as explained in Section~\ref{GSSL}) between the two subsets. A histogram of these scores shows clear separation: sampled images cluster at lower values, while handpicked images have higher surprise scores (mean: $39.45$ for ordinary, $72.35$ for surprising). A Kolmogorov-Smirnov test confirms a significant difference ($D = 0.67, p < 0.001$), supporting our hypothesis.

We collect visual complexity ratings from humans using an online experiment on Prolific \cite{PALAN201822}. Participants are shown pairs of images and asked to select the one that is more visually complex, similarly to the procedure in \cite{saraee2018savoiasdiversemulticategoryvisual}. We followed the sampling strategy in \cite{saraee2018savoiasdiversemulticategoryvisual}, and collect $6000 = 200 \times 30$ comparisons by sampling images with probabilities inversely proportional to the total number of times they have already been selected. We ensure that each pairwise comparison is assessed by three different participants. Hence there are in total $18,000$ evaluations. Each participant evaluates $200$ pairwise comparisons in one session. We also inserted attention checks into the experiment to ensure participant engagement and data reliability. We convert pairwise comparisons into scalar complexity ratings using the Bradley-Terry algorithm \cite{bradley1952rank}. Initially, the final ratings were mostly clustered around zero, a phenomenon also observed in \cite{saraee2018savoiasdiversemulticategoryvisual}. To address this, we applied the same procedure as in their work. Before using the Bradley-Terry model, we rescaled the initial probability matrix from the range $[0,1]$ to $[0.33, 0.66]$. This adjustment helped distribute the final ratings more evenly within a narrower and more interpretable range. We also collect human surprise ratings to validate our LLM-generated scores using a similar experiment.

\subsection{Generating Surprise Scores using LLMs}
\label{GSSL}
In order to have an automated pipeline for predicting complexity, we prompt an LLM using zero-shot Chain-of-Thought  \cite{wei2023chainofthoughtpromptingelicitsreasoning} to get surprise scores for each image in our SVG dataset. Our prompt is shown in Algorithm~\ref{alg:cot}. We selected \gemini for its ability to provide rapid responses without compromising quality. The model generated surprise scores with step-by-step reasoning, producing structured output in the desired format. We also collect human surprise ratings on the SVG image set to validate our LLM-generated scores using a similar Prolific experiment to that described above.

\begin{algorithm}
\caption{Zero-shot-CoT} 
\label{alg:cot}
\vspace{0.1em}
Q: Step by step, explain why this image is surprising or not. Consider factors like rare events, or unexpected content. Be precise in your reasoning. Then, on a precise scale from 0 to 100, rate the surprisal of this image. \\
Provide your reasoning and numeric rating as follows: \\
Reasoning: [your explanation] \\ Rating: $\ll$number$\gg$
\end{algorithm}

\subsection{Multi-Scale Sobel Gradient}
 We propose the Multi-Scale Sobel Gradient (MSG) algorithm (Algorithm~\ref{alg:algo1}) to capture structure in images. MSG applies the Sobel operator across multiple resolutions to RGB images, functioning as an asymmetry detector within $k \times k$ patches, where $k$ represents the kernel size. For symmetric patches, the left and right columns (or rows in horizontal application) of the kernel counterbalance each other, resulting in lower MSG values. Consequently, images exhibiting greater patch-level symmetry produce smaller MSG measurements.  Note that the Sobel operator is typically applied to grayscale images \cite{996}, but we found in ablations that the grayscale version of the algorithm (which first converts the colored images to grayscale) did not perform as well. Based on our permutation tests across 16 datasets, the colored version demonstrated statistically significant superiority over the grayscale version in 6 datasets, while the grayscale version outperformed the colored in 2 datasets. The remaining 8 datasets showed no statistically significant differences between the two versions. 
 Hence we use the color version of the algorithm for the rest of this work.

\begin{algorithm}
\caption{Multi-Scale Sobel Gradient}
\begin{algorithmic}[1]
\REQUIRE Input image $img \in \mathbb{R}^{H \times W \times 3}$ in RGB format
\ENSURE Scalar MSG score
\STATE Normalize image: $img \leftarrow img/255.0$
\STATE Define scales $S = \{1, 2, 4, 8\}$ \\ and weights $W = \{0.4, 0.3, 0.2, 0.1\}$
\STATE Initialize $MSG \leftarrow 0$
\FOR{each scale $s \in S$ and weight $w \in W$}
    \STATE Resize image to $\frac{H}{s} \times \frac{W}{s}$: \\ $scaled \leftarrow Resize(img, (\frac{H}{s}, \frac{W}{s}))$
    \FOR{each channel $c \in \{R,G,B\}$}
        \STATE $G_x = Sobel(scaled[:,:,c], dx=1, dy=0)$
        \STATE $G_y = Sobel(scaled[:,:,c], dx=0, dy=1)$
        \STATE $mag = \sqrt{G_x^2 + G_y^2}$
        \STATE $grad_c \leftarrow Mean(mag)$
    \ENDFOR
    \STATE $s\_grad \leftarrow Mean(grad_R, grad_G, grad_B)$
    \STATE $MSG \leftarrow MSG + w \cdot s\_grad$
\ENDFOR
\RETURN $MSG$
\end{algorithmic}
\label{alg:algo1}
\end{algorithm}

\subsection{Multi-Scale Unique Color}

We introduce MUC which intuitively counts the number of unique colors present in an image. We start from the \enquote{colorfulness} feature introduced by \cite{50×70} (which was first applied to image indexing and content querying) and derive MUC by making colorfulness multi-scale. Note, \cite{50×70} and subsequent works did not provide pseudocode for the colorfulness algorithm - we are the first to do so. MUC has two hyperparameters which control the coarse-graining of spatial resolution and color resolution (bit precision). In the rest of this paper we report results using the best bit precision for each dataset in terms of correlation to complexity (this was typically 7–8 bits). We use a fixed set of spatial scale weights for both MSG and MUC on all datasets.


\begin{algorithm}
\caption{Multi-Scale Unique Color}
\begin{algorithmic}[1]
\REQUIRE Image $I$ (RGB), Number of bits to preserve per channel $b$
\ENSURE Multi-Scale unique color score
\STATE Define scales $S = \{1, 2, 4, 8\}$ \\ and weights $W = \{0.4, 0.3, 0.2, 0.1\}$
\STATE Initialize $MUC \gets 0$
\FOR{each scale $s \in S$ and weight $w \in W$}
\STATE Resize image to $\frac{\text{width}}{s} \times \frac{\text{height}}{s}$: \\ $I_s \gets \text{resize}(I_{rgb}, (\frac{\text{width}}{s}, \frac{\text{height}}{s}))$
\STATE Calculate bit shift: $shift \gets 8 - b$
\STATE Quantize image by bit shifting: \\ $I_q \gets (I_s \gg shift) \ll shift$
\STATE Flatten image: $I_{flat} \gets \text{reshape}(I_q, -1, 3)$
\STATE Create color index: \\ $idx \gets I_{flat}[:, 0] \cdot 2^{16} + I_{flat}[:, 1] \cdot 2^8 + I_{flat}[:, 2]$
\STATE Count unique colors: $n_{\text{unique}} \gets |\text{unique}(idx)|$
\STATE Add weighted contribution: \\ $MUC \gets MUC + (w \cdot n_{\text{unique}})$
\ENDFOR
\RETURN $MUC$
\end{algorithmic}
\end{algorithm}

\section{Results}

\begin{table*}[h]
\begin{center} 
\caption{Model performance on 7 datasets and comparison with previous models. The models from previous work are classified as being based on either handcrafted features or Convolutional Neural Networks (CNNs). * for supervised methods indicate their own \textit{test} set. Bold indicates the best model.\\} 
\label{tab:mainresults} 
\resizebox{1\textwidth}{!}{ 
\renewcommand{\arraystretch}{1.1} 
\begin{tabular}{lccccccccc} 
\hline
\hline
Model/Dataset & \textit{VISC} &  \textit{IC9. Arch.} & \textit{Sav. Art} & \textit{Sav. Suprematism}  & \textit{Sav. Int.} & \textit{IC9. Abstract}  & \textit{SVG}  \\
\hline
\textbf{Handcrafted features} \\
\hline
\quad Corchs 1 (10 features)  & 0.62 & 0.66 & 0.68 & 0.80 & 0.85 & 0.74 & 0.73 \\
\quad Kyle-Davidson 1 (2 features)  & 0.60 &  0.54 & 0.55 & 0.79 & 0.74  & 0.69  & 0.70\\
\hline
\textbf{CNNs} \\
\hline
\quad Saraee (transfer)  & 0.58 & 0.59 & 0.55 & 0.72  & 0.75 & 0.67  & 0.72 \\
\quad Feng (supervised)  & \textbf{0.72} & \textbf{0.92$^\ast$} &  \textbf{0.81} & 0.84  & \textbf{0.89}  & \textbf{0.94$^\ast$}  & 0.83\\
\hline
\textbf{Previous simple model} \\
\hline
\quad $\sqrt{\textit{num\_seg}} + \sqrt{\textit{num\_class}}$ & 0.56  & 0.66 & 0.73 & 0.89 & 0.61  & 0.66 & 0.78 \\
\hline
\textbf{Visual features} \\
\hline
\quad $\textit{MSG}+ \textit{MUC}$ & 0.60  & 0.65 & 0.64 & 0.91 & 0.84  & 0.76 & 0.72 \\
\hline
\textbf{Baseline + visual features} \\
\hline
\quad $\sqrt{\textit{num\_seg}} + \sqrt{\textit{num\_class}} + \textit{MSG}$  & \hspace{0.9cm} 0.68 \qinc{0.13} &   \hspace{0.95cm}0.76 \qinc{0.10} & 0.75 & 0.90  & 0.79  & 0.79 & 0.78 \\
\quad $\sqrt{\textit{num\_seg}} + \sqrt{\textit{num\_class}} +  \textit{MUC}$  & 0.62  & 0.71 & \hspace{0.95cm}\textbf{0.81} \qinc{0.08} & \hspace{0.95cm}\textbf{0.94} \qinc{0.05} &  0.80  & 0.76  & 0.79\\
\quad $\sqrt{\textit{num\_seg}} + \sqrt{\textit{num\_class}} + \textit{MSG}+ \textit{MUC}$  & 0.68 &  0.77 & 0.81 & 0.94  &  \hspace{0.95cm}0.87 \qinc{0.26}  & \hspace{0.89cm} 0.83 \qinc{0.17} & 0.79\\
\hline
\textbf{Baseline + semantic feature} \\
\hline
\quad $\sqrt{\textit{num\_seg}} + \sqrt{\textit{num\_class}} + \textit{Surprise}$  & 0.60 & 0.67& 0.74&  0.89  &  0.60  & 0.67 & \hspace{0.86cm}0.83 \qinc{0.05} \\
\hline
\textbf{Baseline + all features} \\
\hline
\quad $\sqrt{\textit{num\_seg}} + \sqrt{\textit{num\_class}} + \textit{MSG}+ \textit{MUC}+\textit{Surprise}$  & 0.71 & 0.78& \textbf{0.81}&  \textbf{0.94}  &  0.87  & 0.84 & \textbf{0.85} \\
\hline
\hline
\end{tabular} 
}
\label{tab:results}
\end{center} 
\end{table*}

The features we introduced: MSG, MUC, and Surprise, explain additional variance in perceived complexity beyond that captured by segmentation and class counts alone. We present these results in Table~\ref{tab:results}, showing that our models not only surpass all handcrafted features but also achieves performance comparable to the best supervised model. In the following subsections, we present each feature in turn and describe the datasets on which they improve complexity prediction. We then describe our penultimate model, which combines our three novel features to achieve state-of-the-art or near state-of-the-art performance on all datasets.


\vspace{-1em}
\subsection{\center{Explaining the Failure Modes with a Data-Centric Approach: Colors and Structure}}

On the VISC dataset, the MSG feature improves the mean Spearman’s rank correlation coefficient from 0.56 to 0.68, surpassing all previous handcrafted features. Similarly, it achieves a notable increase of 0.10 in mean Spearman’s rank correlation coefficient on the IC9. Architecture dataset. These two datasets comprise of mostly real-world natural images of buildings, and indoor and outdoor scenes which are highly textured or regular - for example a grass field, the windows on a highrise, or rows of seats in an airport. A Sobel operator applied to a spatially coarse-grained image can be interpreted as an (a) symmetry detector. Hence MSG detects (a)symmetry and quantifies the regularity of local patches (e.g. windows on a highrise) at multiple spatial scales, making it a well-suited feature for predicting complexity on VISC and IC9. Architecture.

Table~\ref{tab:results} shows that MUC improves model performance on Sav. Art dataset, increasing Spearman’s correlation from 0.73 to 0.81, and in Sav. Suprematism, where it improves performance from 0.89 to 0.94. These datasets are comprised of 2D art. The efficacy of MUC in these domains likely stems from its ability to quantify color-based complexity, which may be more crucial for people's aesthetic judgments than judgments of naturalistic images.

Two datasets, Sav. Interior Design and IC9. Abstract experience very large performance gains (from 0.61 to 0.87 and from 0.66 to 0.83 respectively) when both MSG and MUC features are included in the model. Interior Design is a dataset of synthetic renderings of rooms for IKEA advertisements. The dataset likely benefits from MSG for the same reason as VISC - the images contain regular, highly symmetric elements like books in a bookshelf or panels of a kitchen cabinet. 
Furthermore, unlike naturalistic images Sav. Interior Design images were designed with color themes in mind for aesthetic reasons, similar to art images which is likely why the Sav. Interior Design dataset benefits from MUC like the Sav. Art dataset.
IC9. Abstract on the other hand, consists of real-life photographs. However, the images are taken in an artistic way with aesthetic value in mind, likely increasing the importance of colorfulness in predicting complexity.

\begin{figure*}
\label{fig1}
    \raggedright
    \begin{minipage}{0.50\linewidth}
        \centering
        \begin{subfigure}{0.47\textwidth}
            \includegraphics[width=\linewidth]{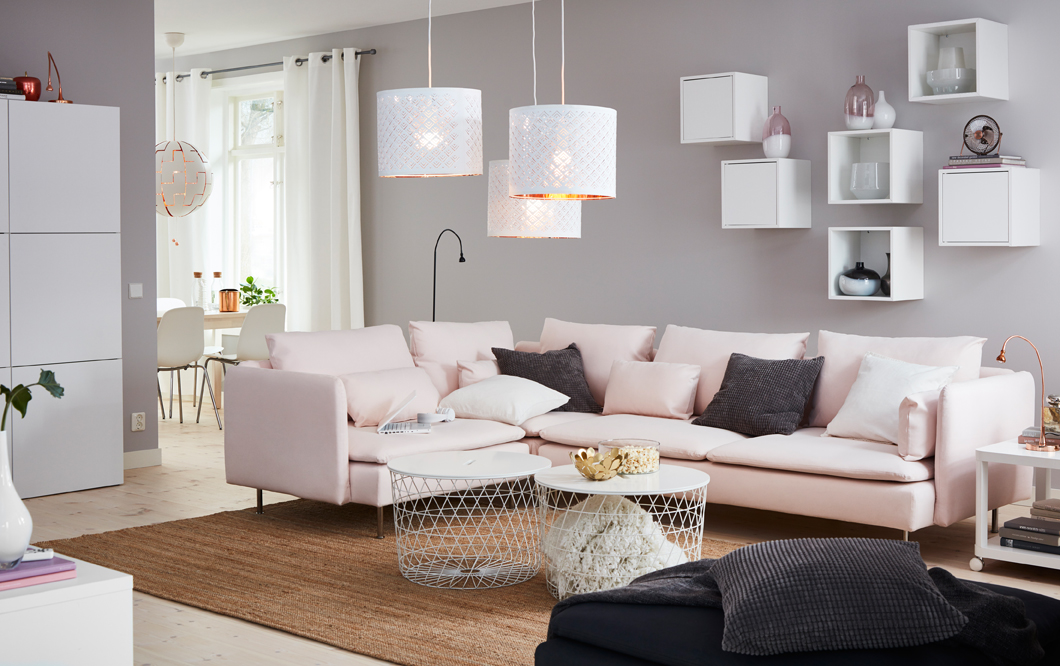}
            \caption{B: 50, P: 39, G: 40}
        \end{subfigure}
        \begin{subfigure}{0.47\textwidth}
            \includegraphics[width=\linewidth]{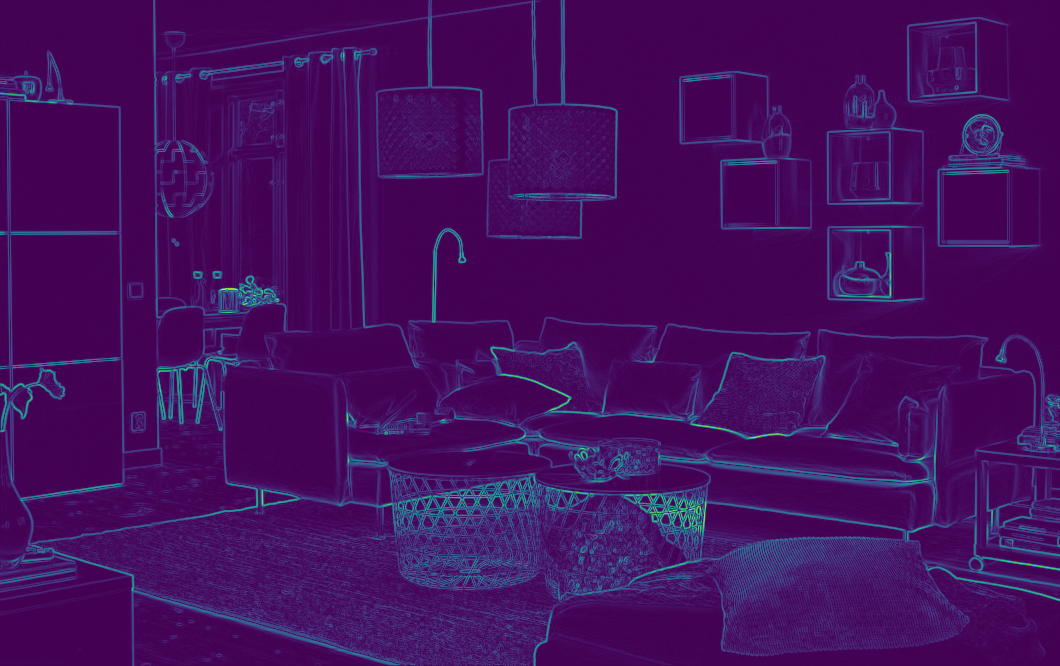}
            \caption{MSG: 20}
        \end{subfigure}
        
        \vspace{0.2cm}
        
        \begin{subfigure}{0.47\textwidth}
            \includegraphics[width=\linewidth]{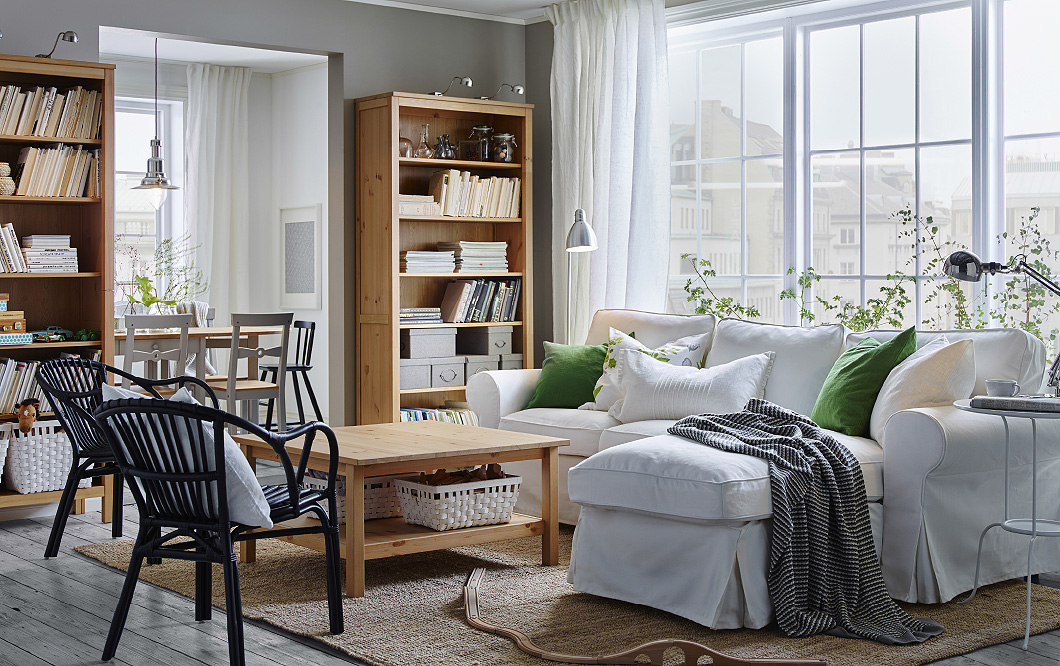}
            \caption{B: 50, P: 67, G: 68}
        \end{subfigure}
        \begin{subfigure}{0.47\textwidth}
            \includegraphics[width=\linewidth]{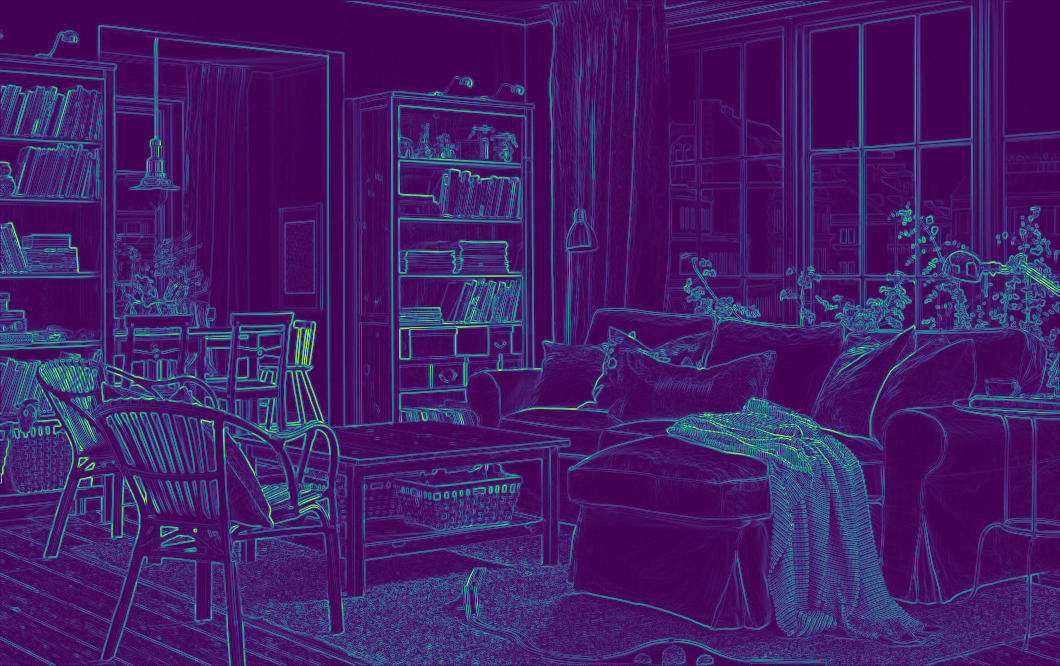}
            \caption{MSG: 92}
        \end{subfigure}

        \caption{Left column: original images from \textit{Sav. Int.}. Right column: gradient visualizations. B: baseline prediction using number of segmentations and classes. G = ground truth complexity. P = predicted complexity using baseline and MSG. All values are scaled between 0 and 100. The first image has 177 segmentations and 35 classes, while the second has 185 segmentations and 38 classes. Due to these similarities, the baseline model predicts nearly identical complexity scores for both. However, MSG acts as a latent dimension, refining predictions to better align with ground truth.}
        \label{fig:examples_fail}
    \end{minipage}
      \hfill
     \begin{minipage}{0.48\linewidth}
       \centering
    \noindent\hrule 
    \vspace{5pt} 
    \begin{subfigure}{0.45\textwidth}
        \includegraphics[width=\linewidth]{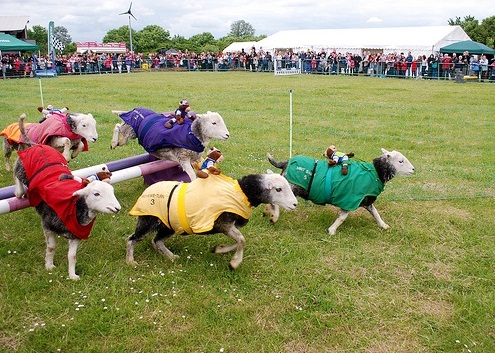}
    \end{subfigure}
    \hspace{0.1cm}
    \begin{subfigure}{0.5\textwidth}
        \caption*{\textit{Surprise score}: 85 \\ \textit{Reasoning}: The image depicts sheep dressed in jockey silks and participating in a race. This is surprising because it is not a typical activity for sheep. The juxtaposition of animals in a human sport is unexpected and humorous.}
    \end{subfigure}

    \vspace{0.1cm}
    
    \begin{subfigure}{0.45\textwidth}
        \includegraphics[width=\linewidth]{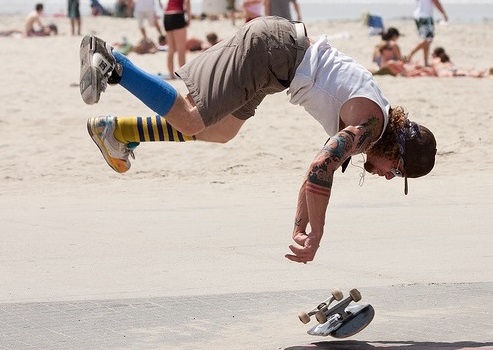}
    \end{subfigure}
    \hspace{0.1cm}
    \begin{subfigure}{0.5\textwidth}
        \raggedright
        \caption*{\textit{Surprise score}: 85\\ \textit{Reasoning}: The image depicts a skateboarder performing an unexpected maneuver on a beach, resulting in an unusual pose mid-air.  The rarity stems from the seemingly uncontrolled backflip and the unusual beach setting for such a stunt.}
    \end{subfigure}
    \vspace{5pt} 
    \hrule 
 \caption{Racing sheep with toy riders or a flying skater from the \textbf{SVG} dataset illustrate the improvement in complexity predictions when incorporating surprise. The baseline model underestimates the visual complexity, with scores of B: 54 and B: 46, while ground truth values are G: 60 and G: 55. Incorporating surprise scores (85) reduces this gap, yielding adjusted predictions of P: 61 and P: 55, demonstrating the role of surprise in aligning predictions more closely with human perception. Explanations provided by \gemini enhance the interpretability of the assigned surprise scores.}
        \label{fig:surp}
    \end{minipage}
    
\end{figure*}

Figure \ref{fig:examples_fail} presents two examples from the Sav. Interior Design dataset where the baseline model \cite{shen2024simplicitycomplexityexplaining} assigns the same complexity due to similar SAM segmentations and FC-CLIP classes, despite differing ground truth complexities. However, the two images differ greatly in their MSG values (20 vs. 92 on a normalized scale of 100), possibly because the lower image has patterns on the bookshelves and blanket which contribute substantially to MSG. The lower (higher) MSG values in the top (bottom) image decrease (increase) the complexity predictions in model, making the predictions much closer to the ground truth complexity ratings.

\subsection{\center{Surprise to the Rescue: Capturing the Complexity Beyond Structural Features}}

Table~\ref{tab:results} shows that including surprise scores generated using \gemini improves Spearman's correlation on the SVG dataset from 0.78 to 0.83 over \cite{shen2024simplicitycomplexityexplaining}'s baseline model. Our model containing surprise scores matches the performance of \cite{feng2023ic9600}'s supervised network. 

\begin{figure}[H]
    \centering
    \includegraphics[width=0.45\textwidth]{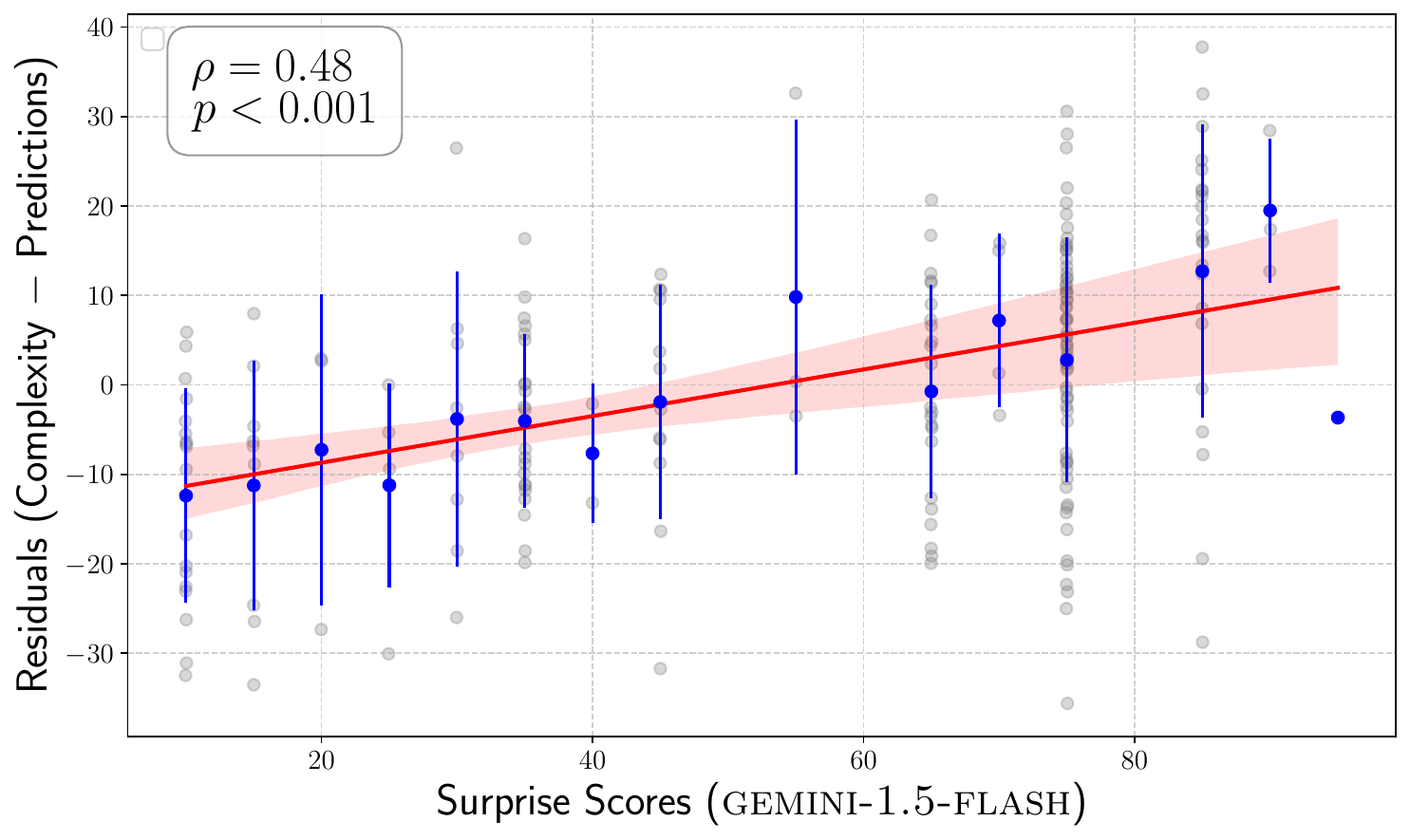}
    \caption{Correlation between residuals (actual complexity - baseline predictions) and surprise scores (SVG).}
    \label{fig:corr}
\end{figure}

Two examples of explanations generated from \gemini using CoT prompting \cite{zhu-etal-2024-explanation, Wei_Jie_2024}, can be seen in Figure~\ref{fig:surp}. In the left image, \gemini successfully identifies the intuitively surprising fact that sheep are racing, an unusual activity for sheep. In the right image, \gemini identifies that it's unusual for the man to be doing an uncontrolled backflip, arguably consistent with human intuition for why the image is surprising.

Furthermore, Figure~\ref{fig:corr} shows that the surprise scores have a correlation of 0.48 ($p< 0.001$) with complexity after regressing out SAM number of segments and FC-CLIP number of objects, showing that surprise captures variance in complexity not explained by the number of segments or objects. These quantitative results are corroborated by participants' anecdotes on how they rated complexity on the SVG dataset. Many reported that \enquote{unusual}, \enquote{surprising}, or \enquote{weird}, objects or events in the image, as well as judgments that the images are \enquote{confusing} or \enquote{unreal} caused them to rate the visual complexity higher. Figure~\ref{box:responses} presents a selection of responses from participants in our experiment, in which they describe the specific strategies they employed to assess visual complexity. Their responses provide a deeper understanding of the cognitive processes involved in evaluating image complexity and highlight how the \textit{element of surprise} works as a latent dimension for visual complexity evaluation.

\begin{figure*}
\begin{center}
\begin{tcolorbox}[
    colframe=gray!50, 
    colback=gray!10, 
    coltitle=black, 
    fonttitle=\bfseries, 
    rounded corners=southwest, 
    arc=6pt, 
    boxrule=0.6pt, 
    width=0.9\textwidth, 
    title={\textbf{Participant Responses}},
    halign=flush left, 
]

\small 

\textbf{Question:} What strategy did you use to rate visual complexity?

\vspace{0.2cm} 
\begin{itemize}[leftmargin=0pt, labelwidth=0pt, labelsep=3pt, itemindent=0pt, itemsep=2pt, parsep=0pt]
    \item \enquote{Checked if they look \textcolor{red}{confusing} or \textcolor{red}{unreal}.}
    \item \enquote{How \textcolor{red}{common} the situation was, the angle the image was taken at, the number of colours, the number of subjects and how in focus they were.}
    \item \enquote{How many different elements there were in each photo and how \textcolor{red}{surprising} or \textcolor{red}{unusual} the images were.}
    \item \enquote{The most complex were those with many different elements to them and the most \textcolor{red}{surprising}.}
    \item \enquote{The \textcolor{red}{weird} or uncanny images appeared quite complex to me, most of the landscape shots or regular street/traffic scenes didn't strike me as complex.}
    \item \enquote{Chose the \textcolor{red}{confusing} ones.}
    \item \enquote{Where there are lots of things to look at. Was there something at a distance in the background? Was the image \textcolor{red}{novel}?}
    \item \enquote{The most complex were the ones with a lot going on in them—so more than one person or something \textcolor{red}{unusual} happening. The least complex were the ones of sport or vehicles which just look like generic images.}
    \item \enquote{If there was a lot going on in the image or if the image was out of the \textcolor{red}{ordinary}. Also, images where there were shadows and different types of light.}
    \item \enquote{I think it was a mix of structure of the image and the elements within it—the more \textcolor{red}{unusual} or juxtaposed the mix, the more complex it appeared to me.}
    \item \enquote{At first, I was thinking about the colours, people, and background. I then thought about how complex the act was in the photo.}
    \item \enquote{I looked at how busy the picture was, then if it wasn't too busy I compared them based on what was happening.}
\end{itemize}

\end{tcolorbox}
\end{center}
\caption{A selection of responses from participants describing the strategies they employed to assess visual complexity. Their responses highlight various factors such as the number of elements, unusual or weird features, level of detail, and how confusing or unreal an image appeared. Words related to surprise, such as ‘unusual,’ ‘confusing,’ and ‘weird,’ have been highlighted in red for emphasis.}
\vspace{-1.5em}
\label{box:responses} 
\end{figure*}

Figure~\ref{fig:surp} shows two examples from the SVG dataset where the model from \cite{shen2024simplicitycomplexityexplaining} underpredicts visual complexity. Both images have high (85) surprise scores. Hence our model including surprise scores correctly increases its complexity predictions, making the predictions much closer to the ground truth complexity. Furthermore, one advantage of using LLMs in our automated pipeline for complexity prediction is LLMs offer (often) interpretable, explicit reasoning in natural language \cite{singh2024rethinkinginterpretabilityeralarge}. Two examples of explanations generated from \gemini using CoT prompting \cite{zhu-etal-2024-explanation, Wei_Jie_2024}, can be seen in Figure~\ref{fig:surp}. In the top image, \gemini successfully identifies the intuitively surprising fact that sheep are racing, an unusual activity for sheep. In the bottom image, \gemini identifies that it's unusual for the man to be doing an uncontrolled backflip, arguably consistent with human intuition for why the image is surprising.

To explore quantitatively how well our \gemini surprise scores aligned with human judgments of surprise, we computed Spearman's correlation between \gemini scores and the human surprise scores we collected. The results revealed a strong positive correlation of 0.73, indicating substantial agreement between model-generated and human-derived assessments of visual surprise. This supports the fact \gemini explanations are good proxies for human judgments of image surprisal which contribute to human perceptions of visual complexity. In summary, \gemini surprise scores improve the predictive power of our complexity model on the SVG dataset and the natural language explanations generated alongside these scores facilitate a clearer understanding of why the image is complex.




The last row of Table~\ref{tab:results} shows the performance of our final model including MSG, MUC, surprise as well as the two features from \cite{shen2024simplicitycomplexityexplaining}. Interestingly, we find combining MSG and MUC with surprise only improves performance on all datasets over using either of these features alone. Furthermore, our final model is significantly more performant than every baseline except \cite{feng2023ic9600}'s supervised network which is competitive. 
Taken together, in contrast to what \cite{shen2024simplicitycomplexityexplaining} previously suggested, the complexity of naturalistic images cannot be explained simply using two generic segment and object features. Some datasets require additional features to explain visual complexity fully.

Lastly, we evaluate the performance of \gemini on our SVG dataset using two distinct prompting strategies.  
In the first approach, the model estimated visual complexity \textit{without} an explicit definition, replicating the open-ended methodology from our Prolific human-subject experiments.  
In the second approach, we provided the canonical definition of visual complexity as \enquote{the level of detail or intricacy contained within an image} \cite{Snodgrass1980} to anchor the model's judgments.  
Both strategies leveraged Chain-of-Thought reasoning to generate structured rationales from \gemini.  
The resulting complexity ratings showed strong internal consistency (Spearman’s \(\rho = 0.89\) between the two strategies) and high alignment with human ground-truth ratings (\(\rho = 0.83\) for both).  
Notably, our final integrated model—which combines pixel-level features, object-level segmentation metrics, and surprise scores from \gemini (quantifying how \textit{surprising} an image is)—outperformed \gemini’s standalone visual complexity evaluations on the SVG dataset.

\begin{figure*}[h]
    \centering
    \begin{tabular}{m{0.5\textwidth} | m{0.4\textwidth}} 
        \begin{minipage}{0.5\textwidth}
            \centering
            \begin{subfigure}[b]{0.49\textwidth} 
                \centering
                \includegraphics[width=\textwidth]{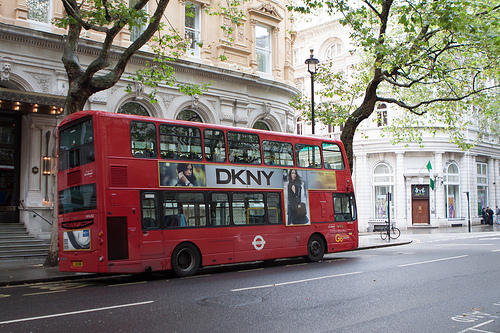}
                \caption{%
                    \centering
                    \begin{tabular}{c}
                        $\textit{num\_seg}$ = 191 \\ 
                        $\textit{num\_class}$ = 9 \\ 
                        \textit{MSG} = 59 \\ 
                        \textit{MUC} = 922 \\ 
                        \underline{\textit{Surprise} = 25}
                    \end{tabular}%
                }
            \end{subfigure}
        \hfill
            \begin{subfigure}[b]{0.49\textwidth} 
                \centering
                \includegraphics[width=\textwidth]{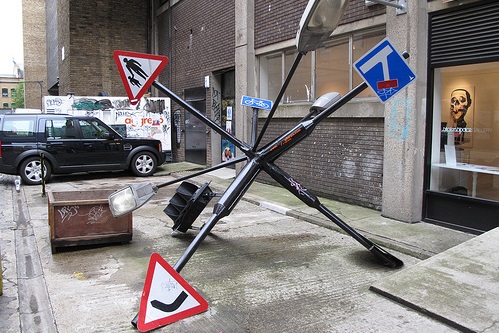}
                \caption{%
                    \centering
                    \begin{tabular}{c}
                        $\textit{num\_seg}$ = 172 \\ 
                        $\textit{num\_class}$ = 9 \\ 
                        \textit{MSG} = 56 \\ 
                        \textit{MUC} = 820 \\ 
                        \underline{\textit{Surprise} = 85}
                    \end{tabular}%
                }
            \end{subfigure}
        \end{minipage} 
        
        & 
        
        \begin{minipage}{0.44\textwidth} 
            \raggedright 
            \vspace{-5em}
            \textbf{Surprise as the hidden dimension} \\
            
            We present a case where both images share similar visual features, including segmentations, classes, MSG, and MUC. The right image, with lower feature values, was initially predicted to have lower complexity. The left image received a complexity score of 60, while the right received 55. However, the element of surprise shifts the outcome, with improved predictions at 52 (72) compared to ground truth values of 25 (74). Despite their similarity, humans may perceive the second image as more complex due to its unusual composition, where multiple traffic signs are clustered together.
            
        \end{minipage}
    \end{tabular}
    \captionsetup{justification=centering, margin=2cm}
    \caption{Comparison of two images having similar values of visual features but with differing complexity evaluations.}
    \label{fig:image_comparison}
\end{figure*}

In Figure~\ref{fig:image_comparison}, we present another pair of images where, this time, the addition of surprise scores causes their predicted complexity values to diverge in opposite directions.

\section{Discussion} 

This work introduces three features to solve the inadequacy of the generic constructs proposed by \cite{shen2024simplicitycomplexityexplaining} to explain complexity in certain datasets: multi-scale gradient analysis, multi-scale unique color, and a surprise score. We experimentally show that complexity can be multifaceted - requiring both generic features (segmentation and object counts) and dataset specific features to predict accurately. Our results demonstrate that pixel-level structural and chromatic features, MSG and MUC effectively capture key aspects of visual complexity, with MSG quantifying continuous spatial intensity variations across scales and MUC capturing color diversity, both complementing object-level information from deep segmentation models. Additionally, we introduce surprise, an image-level \enquote{cognitive} feature, as a previously underexplored dimension of visual complexity, representing the degree to which the image as a whole deviates from expected objects or events. We introduce a new dataset SVG containing images with unexpected or novel elements to show that surprise contributes significantly to perceived complexity. We constructed our SVG dataset by manually selecting surprising images and hence acknowledge the limitations in scalability and the potential subjectivity of this approach. We hope our preliminary yet confirmatory results will inspire future research to develop larger datasets using objectively defined measure to further validate the link between surprise and complexity.

\begin{figure*}[h]
  \centering
  \captionsetup[subfigure]{labelformat=empty,font=scriptsize}

  \begin{tabular}{p{0.14\textwidth}p{0.14\textwidth}p{0.14\textwidth}p{0.16\textwidth}p{0.14\textwidth}p{0.14\textwidth}}
    \centering{\textit{VISC}} & \centering{\textit{IC9. Architecture}} & \centering{\textit{Sav. Art}} & 
    \centering{\textit{Sav. Suprematism}} & \centering{\textit{Sav. Int. Design}} & \centering{\textit{IC9. Abstract}} \tabularnewline
    \end{tabular}

  \begin{subfigure}{0.16\textwidth}
    \includegraphics[width=\linewidth,height=\linewidth]{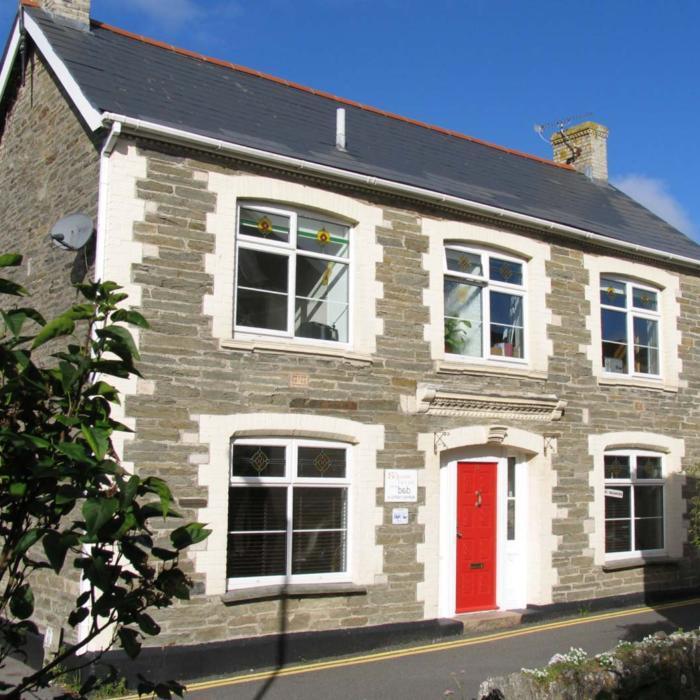}
    \caption{\footnotesize{B: 52},
             P: 62,
             G: 63
            }
  \end{subfigure}
  \hfill
  \begin{subfigure}{0.16\textwidth}
    \includegraphics[width=\linewidth,height=\linewidth]{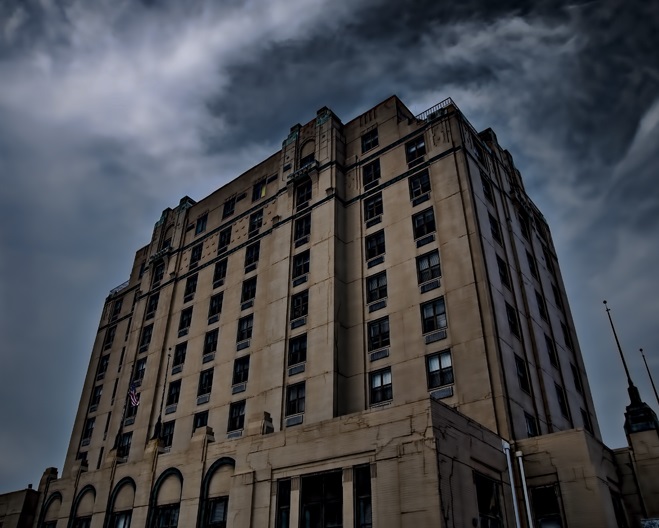}
    \caption{\footnotesize{B: 52},
             P: 44,
             G: 44}

  \end{subfigure}
  \hfill
  \begin{subfigure}{0.16\textwidth}
    \includegraphics[width=\linewidth,height=\linewidth]{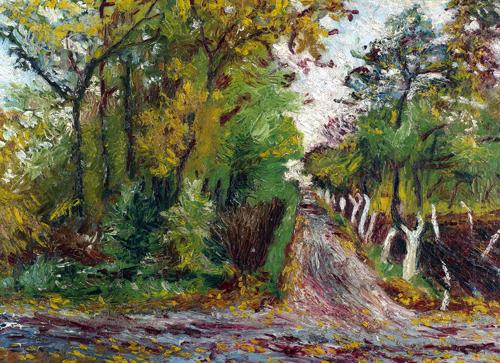}
    \caption{\footnotesize{B: 52},
             P: 66,
             G: 68}
  \end{subfigure}
  \hfill
  \begin{subfigure}{0.16\textwidth}
    \includegraphics[width=\linewidth,height=\linewidth]{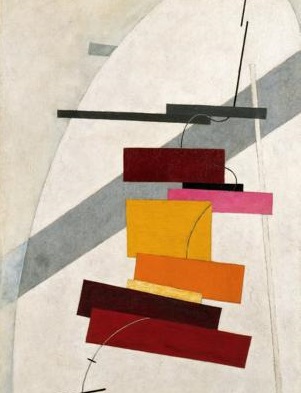}
    \caption{\footnotesize{B: 52},
             P: 43,
             G: 41}
  \end{subfigure}
  \hfill
  \begin{subfigure}{0.16\textwidth}
    \includegraphics[width=\linewidth,height=\linewidth]{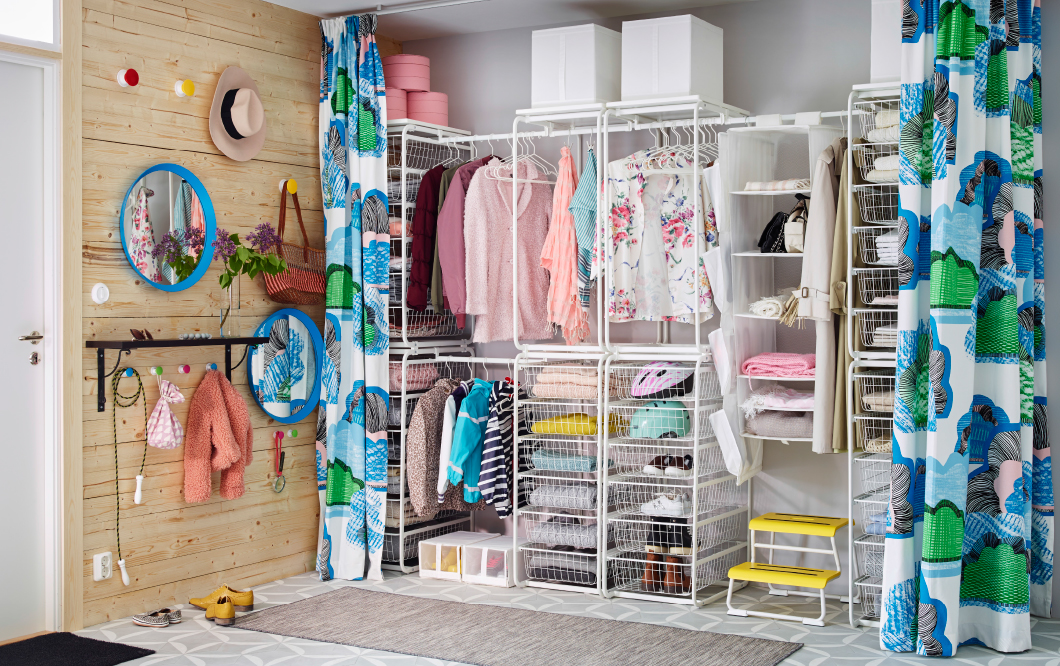}
    \caption{\footnotesize{B: 52},
             P: 92,
             G: 94}
  \end{subfigure}
  \hfill
  \begin{subfigure}{0.16\textwidth}
    \includegraphics[width=\linewidth,height=\linewidth]{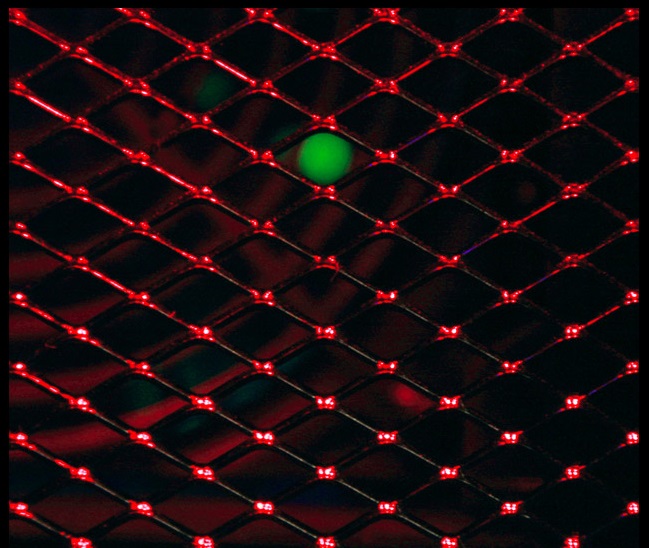}
    \caption{\footnotesize{B: 52},
             P: 38,
             G: 38}
  \end{subfigure}

  \vspace{2mm}

  \vspace{-0.2cm}
  \caption{Representative samples from previous datasets show initial complexity predictions ($B = 52$), derived from $\sqrt{\textit{num\_seg}}$ and $\sqrt{\textit{num\_class}}$, refined using \textbf{MSG} (VISC, IC9 Architecture), \textbf{MUC} (Sav. Art, Sav. Suprematism), and \textbf{MSG + MUC} (Sav. Int. Design, IC9 Abstract). The enhanced predictions ($P$) align closer to ground truth scores ($G$).}
  \label{fig:examples}
\end{figure*}

\vspace{-1em}
\paragraph{Numerosity Perception}
As the baseline approach, \cite{shen2024simplicitycomplexityexplaining} uses the square root of the number of segmentations and classes, an empirical finding that demonstrated better performance than using the raw values. One reason why the square root works better than the raw values can be explained by the Weber-Fechner Law \cite{moyer1967time}, which establishes that subjective perception typically relates to stimulus intensity in a \textit{nonlinear} manner. According to this principle, the just-noticeable difference between two stimulus intensities is proportional to the absolute stimulus intensity. From a practical perspective, this means participants can more readily distinguish between images containing 10 versus 20 segmentations than between those with 110 versus 120 segmentations, despite the absolute difference being identical. We also tested applying logarithmic transformation to the number of segmentations and classes, but the square-root transformation achieved slightly better results in the regression model, so we decided to stick with it.

\vspace{-1.4em}
\paragraph{MSG versus Patch Symmetry and Edge Density}

Patch symmetry \cite{KYLEDAVIDSON2023105319, Olivia2004IdentifyingTP, Rosenholtz2007MeasuringVC} and edge density \cite{GUO2018110, 10.1007/978-3-319-23222-5_15} are two commonly used low-level features for capturing structural information in an image. We run ablation studies to investigate whether MSG is more performant than these two features. We evaluate on all 16 datasets: VISC, RSIVL, SAVOIAS (5 categories - excluding Advertisement and Visualizations due to heavy presence of text), IC9600 (8 categories), and SVG using permutation tests. We compare MSG against patch symmetry because \cite{shen2024simplicitycomplexityexplaining} reported that it captures structural information well on the VISC dataset. We compare MSG against Canny edge density because Canny is a standard metric for predicting visual complexity. It has been applied in \cite{GUO2018110, corchs2016predicting, Rosenholtz2007MeasuringVC}.
MSG performs significantly better than Canny edge density in five dataset (IC9. Abstract, IC9. Paintings, Sav. Scenes, IC9. Advertisement, and Sav. Interior Design). For example, on IC9. Abstract, MSG achieves significantly higher correlation with the complexity compared to edge density (0.517 vs 0.705). Edge density only outperformed MSG on two datasets (IC9. Person and Sav. Art). Compared to patch symmetry, MSG performs better on 7 datasets (IC9. Abstract, IC9. Paintings, IC9. Scenes, IC9. Advertisement, IC9. Architecture, IC9. Person, and Sav. Art). Patch symmetry outperforms MSG on only one dataset (Sav. Interior Design dataset). Our ablations show that MSG is currently the most performant structural feature for predicting complexity, outperforming both edge density and patch symmetry across diverse image datasets.
\vspace{-1em}

\paragraph{Single versus Multi-Scale} We perform ablation studies to investigate the necessity of multi-scale computations for MSG and MUC. We measure the correlation between complexity and our visual features using both single and multiple scales. By single scale, we refer to the application of our algorithms on the whole image only once ($W = {1.0}$ and $S = {1}$) without rescaling. We find that multi-scale MSG outperforms single-scale MSG on 8 datasets  (VISC, IC9 Abstract, Sav. Objects, Sav. Scenes, Sav. Suprematism, IC9 Transport, IC9 Objects, and SVG), while single-scale MSG outperforms multi-scale MSG on just two datasets (IC9 Paintings and IC9 Architecture) out of the 16 datasets tested. The advantage of the multi-scale approach was particularly pronounced in Sav. Suprematism, where correlation increased substantially from 0.474 to 0.616. 
Similarly, multi-scale MUC outperformed single-scale MUC on 5 datasets (IC9. Abstract, IC9. Paintings, Sav. Art, IC9. Advertisement, IC9. Architecture), while single-scale MUC outperformed multi-scale MUC on only one dataset (IC9 Transport). The multi-scale versions of MSG and MUC maximize the correlation with the perceived visual complexity.

\vspace{-1em}
\paragraph{Memorability and Aesthetics} 
The existing literature on visual complexity has largely concentrated on identifying low-level image features that define complexity, rather than examining its perceptual implications for the observer. However, as noted by \cite{10.1007/978-3-642-02728-4_17}, complexity inherently relates to the difficulty of cognitive processing, suggesting that surprising images should be perceived as more complex. We provide evidence in support of this hypothesis, showing that surprise, as determined by an LLM, contributes significantly to visual complexity on our SVG dataset. \\
Prior work links image memorability, a closely related concept to complexity \cite{kyle2023complexity}, to distinctiveness, finding that unique or atypical features enhance retention. \cite{Bruce1994, Bartlett1984} find that distinct facial features are more memorable. \cite{Lukavsky2017} find that images that deviate from their neighbors in a CNN embedding space have increased recall.
We observe a weak positive correlation between human surprisal ratings and both AMNet-predicted 
($\rho = 0.25, p < 0.05$) \cite{fajtl2018amnetmemorabilityestimationattention} and 
ResMem-predicted memorability ($\rho = 0.25, p < 0.05$) \cite{needell2022embracingnewtechniquesdeep}, suggesting that surprise may contribute to, or even be a common cause for both judgments of complexity and memorability. Future works should elucidate the relationship between complexity, memorability, and surprise. \\
The relationship between complexity and aesthetics has been extensively studied, though its precise functional form remains uncertain. \cite{IMAMOGLU20005, Berlyne1970} provides evidence for an inverted U-shape between visual complexity and aesthetics. \cite{10.1007/978-3-319-16178-5_2} confirms the ascending part of Berlyne’s inverted-U curve and employs complexity features to predict beauty judgments in images. However, their measure relies solely on visual features like edges and symmetry, while the authors acknowledge that semantic factors, such as familiarity, may also shape perceived visual complexity. In this work, we find preliminary evidence that surprise contributes to complexity. Given the close relationship between complexity and aesthetics, it would be interesting in future works to investigate the extent to which surprise also influences aesthetic, potentially by mediating complexity. Our findings emphasize the multifaceted nature of complexity, requiring varied features across datasets. An incomplete representation of visual complexity may obscure its link to aesthetics, while a more comprehensive approach integrating visual and cognitive-semantic features could enhance understanding.
\vspace{-1.0em}
\paragraph{Future work} 

While our work uses whole-image, holistic surprise scores, \gemini identifies specific object relationships and object-contexts that make each image more or less surprising. Therefore, it would be interesting for future works to derive surprise scores (or even other semantic features) from object-centric representations such as scene graphs. Our dataset, SVG enables this type of investigation by including human-rated complexity scores, and inheriting region descriptions, and scene graphs from the Visual Genome dataset. These rich annotations open a promising path for developing and validating features that reflect semantic rather than purely visual contributions to complexity.

\bibliographystyle{apacite}

\setlength{\bibleftmargin}{.125in}
\setlength{\bibindent}{-\bibleftmargin}
\bibliography{CogSci_Template}

\section{Appendix}

\subsection{Results on All Datasets}
\label{ap:all_datasets}

Table~\ref{tab:comparison} compares the performance of the baseline approach, our final model, and Feng's supervised model across all datasets.

\begin{table*}[h]
\begin{minipage}{0.5\textwidth} 
\begin{center} 
\caption{Correlations with the ground truth perceived complexity ratings. Bold indicates the best model performance for each dataset. Improvements from the baseline are indicated with \textcolor{IncrGreen}{arrows}.} 
\label{tab:comparison} 
\huge 
\resizebox{\textwidth}{!}{ 
\renewcommand{\arraystretch}{1.05} 
\begin{tabular}{l|c|c|c} 
\hline
\hline
Dataset & Baseline & \textit{Our final model} & Feng (sup.) \\
\hline
VISC &  0.56 & 0.71 \qinc{0.15} & \textbf{0.72} \\
\hline
RSIVL &  0.83 &  \textbf{0.84} \qinc{0.01} &  0.83 \\
\hline
SVG & 0.78 & \textbf{0.85} \qinc{0.07} & 0.83 \\
\hline
Sav. Art &  0.73 & \textbf{0.82} \qinc{0.09} & \textbf{0.82} \\
Sav. Sup. & 0.89 & \textbf{0.94} \qinc{0.05} & 0.84 \\
Sav. Int. & 0.61  & 0.87  \qinc{0.26} & \textbf{0.89} \\
Sav. Objects & 0.82 &  \textbf{0.85} \qinc{0.03} & 0.83 \\
Sav. Scenes & 0.77 & 0.81 \qinc{0.04} & \textbf{0.87} \\
\hline
IC9. Arch. & 0.66 & 0.77 \qinc{0.11} & \textbf{0.93} \\
IC9. Abstract & 0.66  & 0.84 \qinc{0.18} & \textbf{0.95} \\
IC9. Scenes & 0.77 & 0.83 \qinc{0.06} & \textbf{0.93} \\
IC9. Objects & 0.80 &  0.84 \qinc{0.04} & \textbf{0.92} \\
IC9. Transport & 0.74 & 0.79 \qinc{0.05} & \textbf{0.88} \\
IC9. Paintings & 0.83 & 0.87 \qinc{0.04} & \textbf{0.95} \\
IC9. Adv. & 0.75 &  0.82 \qinc{0.07} &  \textbf{0.93} \\
IC9. Person & 0.59 &  0.66 \qinc{0.07} & \textbf{0.85} \\
\hline
\hline
\end{tabular} 
}
\end{center}
\end{minipage}
\hfill
\begin{minipage}{0.47\textwidth} 
\raggedright
Here we compare the baseline approach 
($\sqrt{\textit{num\_seg}} + \sqrt{\textit{num\_class}}$), our final model ($\text{baseline} + \textit{MSG}+ \textit{MUC}+\textit{Surprise}$), and the best supervised model \cite{feng2023ic9600} across 16 datasets. 

When evaluating on datasets not included in Feng's training data, our model outperforms Feng's supervised approach on 4 out of 8 datasets, while Feng's model performs better on 3 datasets (with one dataset resulting in a tie), giving us a final record of 4/3/1.

Feng's supervised method shows superior performance across all IC9 datasets as expected since their model was specifically trained on this data distribution (in-domain evaluation). In contrast, our approach maintains strong performance while using interpretable features that don't require training on any dataset.
\end{minipage}
\end{table*}

\subsection{Additional Experiment Details}
\label{ap:experiment}

In our main paper, we briefly described the creation of our Surprising Visual Genome (SVG) dataset and our methodology for collecting human complexity ratings. Here, we provide additional details about our experimental procedure.

The visual complexity assessment task was conducted using Prolific \cite{PALAN201822} as the participant recruitment platform. Each participant completed 200 pairwise comparisons in a single session. The study, which took approximately 15 minutes to complete, offered payment of 2.61 pounds. All participants were recruited from the UK.

\begin{figure*}[h]
    \centering
    \includegraphics[width=1.0\linewidth]{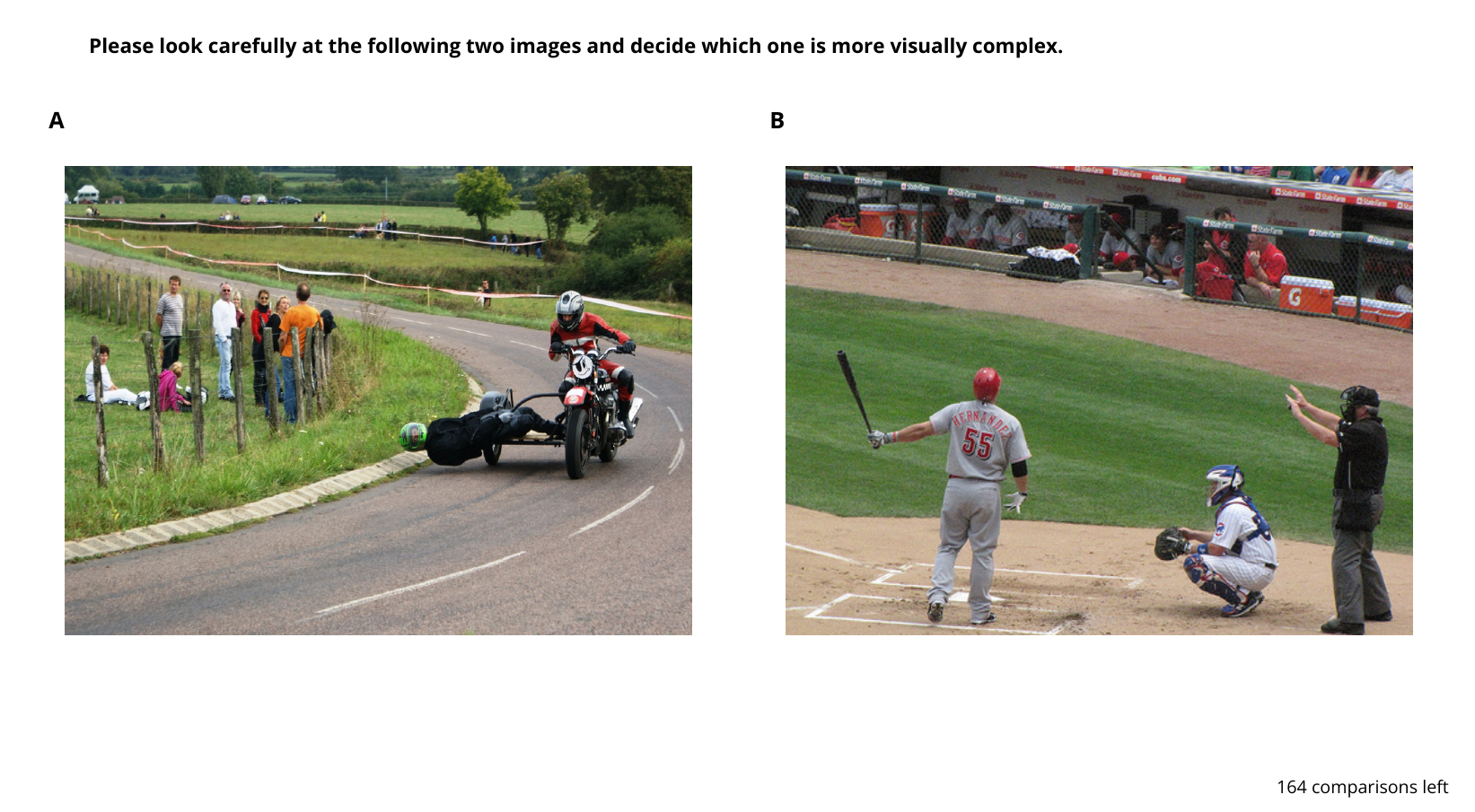}
    \caption{Interface of the pairwise comparison experiment for visual complexity assessment. Participants were instructed to select the image they perceived as more visually complex by clicking on it. The number of comparisons remaining was displayed in the bottom right corner.}
    \label{fig:experiment_appendix}
\end{figure*}

The experiment was implemented using jsPsych \cite{deLeeuw2015jsPsych}, a JavaScript library for creating behavioral experiments in web browsers. After reading instructions and completing practice trials, participants were presented with image pairs and asked to select the one they perceived as more visually complex (see Figure~\ref{fig:experiment_appendix}). Initial instructions are provided in Frame~\ref{fig:instructions}.

\begin{figure*}[t]  
\begin{framed}
\begin{minipage}{0.9\textwidth} 
\vspace{0.1cm} 
Thank you for your consent to participate in our study.
In this study, you will be shown pairs of images and asked to judge which one appears more \textbf{visually complex} to you.

You will make around 200 comparisons in total. For each pair, simply click on the image you judge to be more surprising. There are no right or wrong answers - we are interested in your perception of surprise.

At the end of the study, we will ask brief questions to better understand your judgements of visual complexity.
The complete study will take approximately 15 minutes.

\textbf{We will now show you some example pairs to familiarize you with the task. Click \enquote{Start} to begin.}
\vspace{0.1cm} 
\end{minipage}
\end{framed}
\caption{Initial instructions provided to participants at the beginning of the visual complexity experiment}
\label{fig:instructions}
\end{figure*}

To ensure data quality, we embedded three attention check trials among the 200 comparisons. In these trials, participants were shown identical images, except one image contained a text overlay with explicit instructions to select it. Participants who failed more than one attention check were excluded from the analysis and replaced with new participants to maintain our target sample size.

After completing all comparisons, participants responded to questions about their strategy for rating visual complexity, whether they found particular types of images consistently more complex, and if they had additional comments about their experience. These responses provided qualitative insights into the factors influencing complexity judgments, as highlighted in the main paper.

The pairwise comparison data was processed using the Bradley-Terry algorithm \cite{bradley1952rank} to convert the binary choices into continuous complexity scores. To address the issue of scores clustering around zero, we applied the rescaling procedure described in \cite{saraee2018savoiasdiversemulticategoryvisual}, which involved transforming the probability matrix from [0,1] to [0.33, 0.66] before applying the Bradley-Terry model.

To collect human surprise ratings, we utilized the same experimental setting with minor modifications to the instructions and text to focus on surprisal rather than complexity.

Our complete experimental materials, including stimuli, code for the web-based interface, data processing scripts, and analysis pipelines, are available at \url{https://github.com/Complexity-Project/Experiment} to facilitate replication and extension of our work.

\subsection{SVG dataset}
\label{ap:svg}

\begin{figure*}[h]
  \centering
  \captionsetup[subfigure]{labelformat=empty,font=scriptsize}

  \begin{tabular}{p{0.125\textwidth}p{0.17\textwidth}p{0.10\textwidth}p{0.16\textwidth}p{0.13\textwidth}p{0.14\textwidth}}
\tabularnewline
    \end{tabular}

  \begin{subfigure}{0.16\textwidth}
    \includegraphics[width=\linewidth,height=\linewidth]{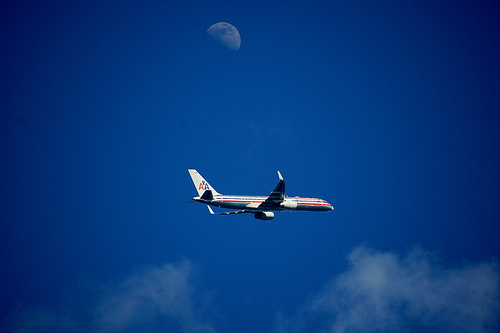}
    \caption{\scriptsize{G: 15,
             H: 4,
             C: 5}
            }
  \end{subfigure}
  \hfill
  \begin{subfigure}{0.16\textwidth}
    \includegraphics[width=\linewidth,height=\linewidth]{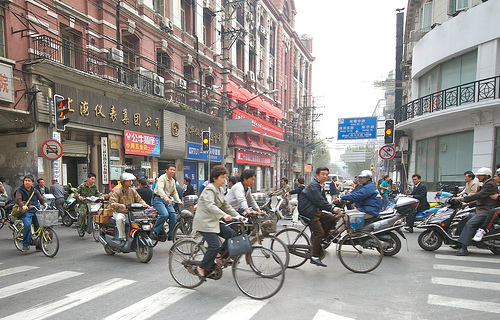}
    \caption{\scriptsize{G: 35,
             H: 32,
             C: 81}
            }
  \end{subfigure}
  \hfill
  \begin{subfigure}{0.16\textwidth}
    \includegraphics[width=\linewidth,height=\linewidth]{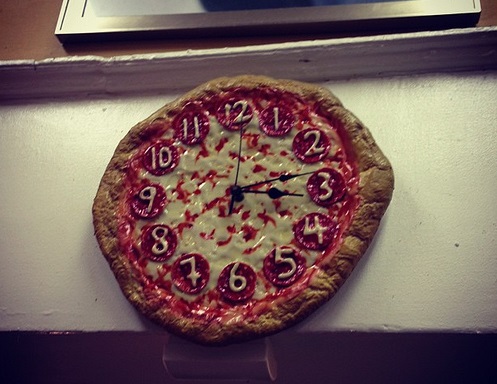}
    \caption{\scriptsize{G: 65,
             H: 66,
             C: 27}
            }
  \end{subfigure}
  \hfill
  \begin{subfigure}{0.16\textwidth}
    \includegraphics[width=\linewidth,height=\linewidth]{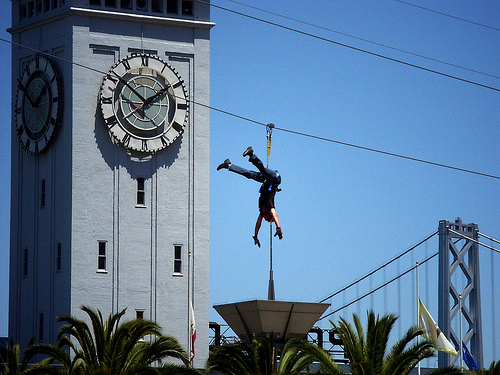}
     \caption{\scriptsize{G: 75,
             H: 76,
             C: 56}
            }
  \end{subfigure}
  \hfill
  \begin{subfigure}{0.16\textwidth}
    \includegraphics[width=\linewidth,height=\linewidth]{figures/2333800.jpg}
    \caption{\scriptsize{G: 85,
             H: 98,
             C: 72}
            }
  \end{subfigure}
    \hfill
  \begin{subfigure}{0.16\textwidth}
    \includegraphics[width=\linewidth,height=\linewidth]{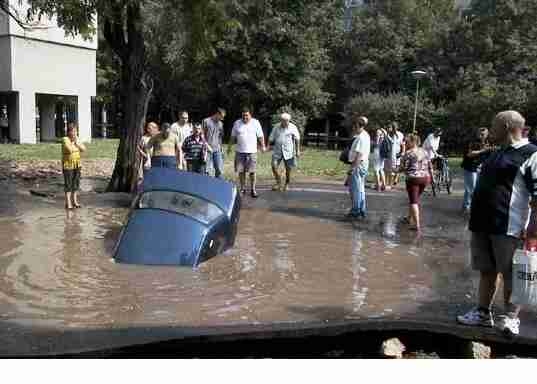}
    \caption{\scriptsize{G: 85,
             H: 85,
             C: 81}
            }
    
  \end{subfigure}

  \vspace{2mm}

  \vspace{-0.2cm}
  \caption{Representative samples from the SVG dataset. G indicates the \gemini-generated surprise scores, H represents human surprise scores collected through our experiments, and C shows the ground truth perceived visual complexity - all values are normalized to a 0-100 scale. The second figure illustrates cases where relatively ordinary images tend to be classified as visually complex, due to other dimensions of complexity such as high number of objects or segmentations. Figures 4 and 5 share similar visual features (number of segmentations = 88 vs. 88, number of classes = 5 vs. 4, MSG = 37 vs. 38, MUC = 34 vs. 24), yet their visual complexity values differ significantly (56 vs. 72). This difference may be attributed to how participants perceived the latter image as more surprising (76 vs. 98).}
  \label{fig:svg_examples}
\end{figure*}

To curate the SVG dataset, we employed a targeted approach, first manually identifying candidates with high surprisal value from the Visual Genome dataset, then validating our selections using an automated method (using an LLM). We considered an alternative strategy—directly evaluating surprise values across the entire corpus using an LLM and selecting those exceeding a threshold—but determined this was impractical for the following: Identifying truly distinctive surprising images, as exemplified in Figure~\ref{fig:svg_examples} and subsequently confirmed by participant responses in Figure~\ref{box:responses}, required examining a substantial portion of the dataset, which consists of $\geq 100$K images. The computational demands of LLM processing at this scale imposed substantial practical constraints: either (1) lengthy processing times due to API rate limits, or (2) prohibitive costs for premium API access that would bypass these limitations. Therefore, we implemented a hybrid approach, first manually selecting candidate images with high surprisal potential, followed by LLM verification to ensure selection validity.

\subsection{Ablations on LLMs for Surprise Scores}
\label{ap:llm_ablation}
To generate surprise scores for images in our SVG dataset, we evaluated three different LLMs: \gemini \cite{geminiteam2024gemini15unlockingmultimodal}, \llama \cite{grattafiori2024llama3herdmodels}, and \chatgpt \cite{openai2024gpt4technicalreport}. Our selection strategy aimed to compare models across the accessibility spectrum: \llama as a fully open-source model with unrestricted access, \chatgpt as a sophisticated commercial model available only through paid API (or free account using the UI), and \gemini as a middle-ground solution offering free access with daily request limitations.

We input the exact same Chain-of-Thought prompt (as given in Algorithm~\ref{alg:cot}) to the models.
Although a popular choice as being fully open-source, \llama could not keep up with the other models, as it struggled with understanding the task and providing responses that could be parsed by our algorithm. Therefore, we decided to focus on comparing \gemini and \chatgpt for our task. In Table~\ref{tab:llm_ablation}, we present the Spearman correlation coefficients between our predicted values and ground truth perceived visual complexity, comparing surprise scores derived from \gemini, \chatgpt, and human participants as features in our regression models.

\bgroup
\def\arraystretch{1.4}
\begin{table*}[ht]
\captionsetup{justification=centering}
\caption{Correlations of our models with the ground truth perceived visual complexity.}
\centering

\begin{tabular}{c|ccc}
\toprule
     & \multicolumn{3}{c}{\textbf{\begin{tabular}[c]{@{}c@{}}Surprise scores\end{tabular}}}
     \\
\textbf{Other features included}

& \multicolumn{1}{c}{\chatgpt}             & \multicolumn{1}{c}{\gemini}  & \multicolumn{1}{c}{Human-rated}  \\\hline
\begin{tabular}{@{}c@{}}
$\sqrt{\textit{num\_seg}} + \sqrt{\textit{num\_class}}$ 
\end{tabular}
& 
0.81
& 

0.83
& 

\textbf{0.85}
\\\hline
\begin{tabular}{@{}c@{}}
$\sqrt{\textit{num\_seg}} + \sqrt{\textit{num\_class}} + \text{MSG} + \text{MUC}$ 
\end{tabular}
&               
\begin{tabular}{@{}c@{}}
0.82
\end{tabular}  
& 
\begin{tabular}{@{}c@{}}
0.85
\end{tabular}
&
\textbf{0.86}

\\\hline

\bottomrule
\end{tabular}
\label{tab:llm_ablation}
\end{table*}
\egroup

As anticipated, human-rated surprise scores yield the strongest contribution to the baseline model, enhancing correlation from 0.78 to 0.86. \gemini and \chatgpt-generated surprisal scores show substantial agreement with a correlation of 0.78. However, the \gemini-generated surprise scores demonstrate superior performance compared to those produced by \chatgpt, with the latter failing to surpass Feng's supervised model (0.83). Based on these findings, we ultimately selected \gemini-produced surprise scores for our final model—a choice that offered the additional practical advantage of free access within certain usage limits.

\subsection{Further Details on MSG and MUC}
\label{ap:ablations}

\begin{figure*}[ht]
    \begin{minipage}{0.48\linewidth}
        \centering
        \includegraphics[width=\linewidth]{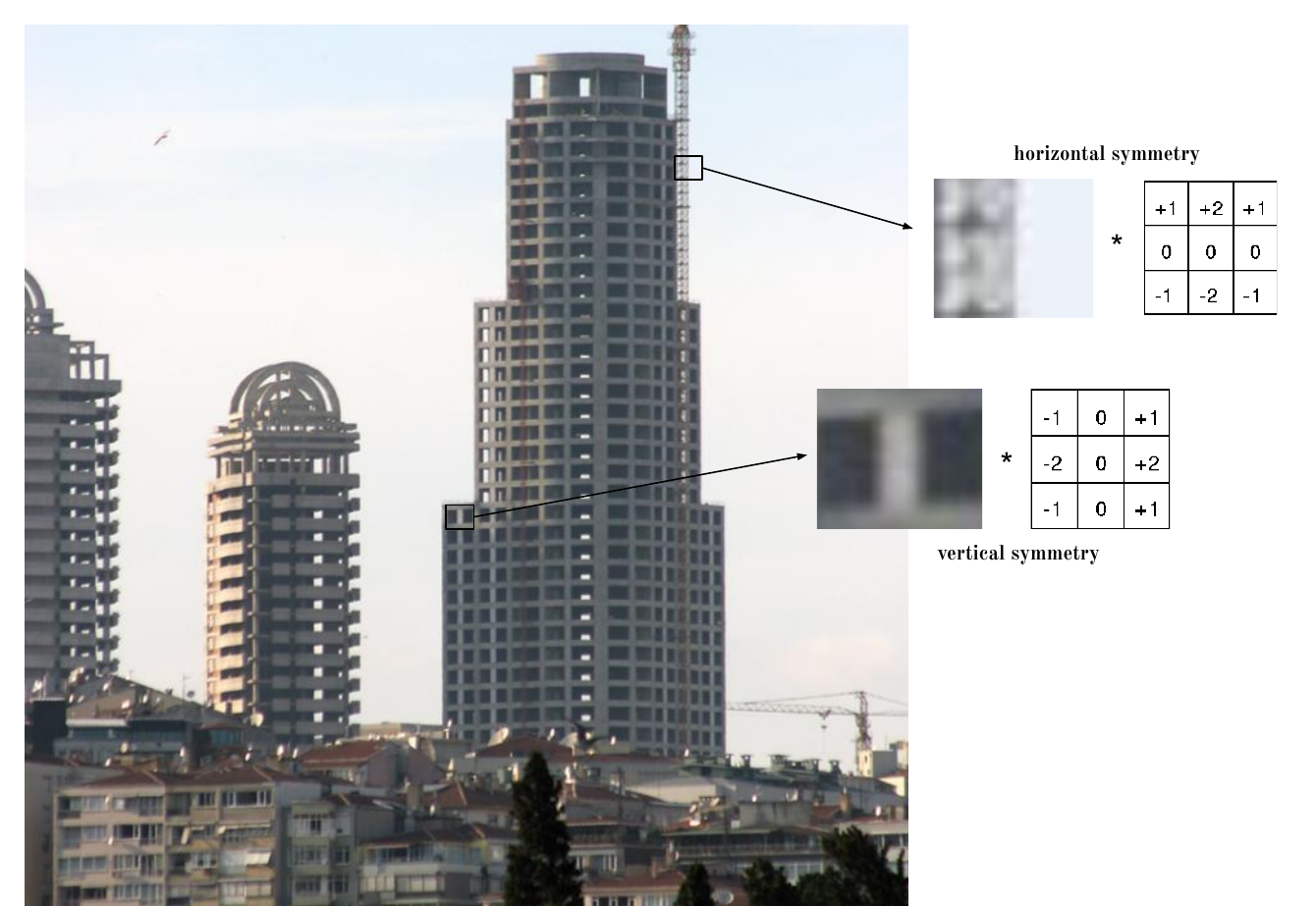}
    \end{minipage}%
    \begin{minipage}{0.48\linewidth}
        MSG applies the Sobel operator to each $k \times k$ patch in the image. While the Sobel operator has symmetric magnitude, its kernel contains opposite signs on different sides. Consequently, symmetrical image patches typically produce lower MSG values. This image has a high segmentation count (538 - maximum in VISC dataset), yielding a high baseline prediction of 86 versus the ground truth of 56. Incorporating its MSG score of 0.26 (50th percentile) reduces the predicted complexity to 0.69, aligning closer to the actual value.
    \end{minipage}
    
    \vspace{1em}
    
    \begin{minipage}{0.48\linewidth}
        \centering
        \includegraphics[width=\linewidth]{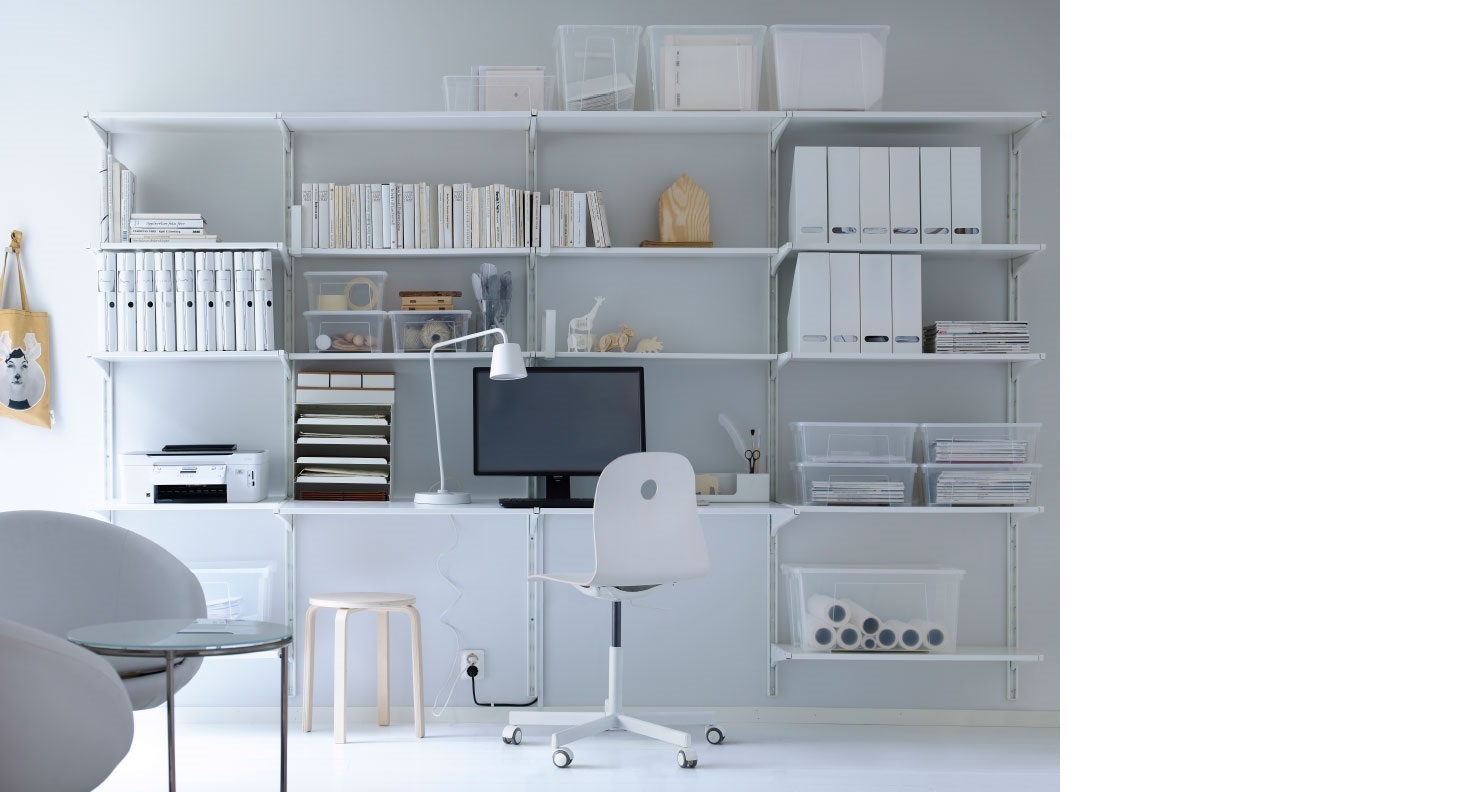}
    \end{minipage}%
    \begin{minipage}{0.48\linewidth}
        Similarly, this image receives a high baseline prediction (57) due to its substantial segmentation count (204, 78.5 percentile) and class diversity (37, 87 percentile). However, by incorporating both symmetry effects and color uniformity, we can refine our prediction. The image's low MSG value (0.19, 15th percentile) and MUC value (0.18 for 7 bits, 7th percentile) adjust the final prediction to 0.20, much closer to the ground truth complexity of 0.16. Both the symmetrical patches and monochromatic structure reduce its perceived visual complexity.
    \end{minipage}
    
    \caption{Examples where MSG and MUC features help improve complexity predictions: (top) an image from VISC dataset and (bottom) an image from Sav. Int. dataset, both previously shared by \cite{shen2024simplicitycomplexityexplaining} as failure modes.}
    \label{fig:msg_muc_examples}
\end{figure*}

Here we present two examples in Figure~\ref{fig:msg_muc_examples} from the previous study \cite{shen2024simplicitycomplexityexplaining}, presented as failure modes.

We performed multiple ablation studies across all 16 datasets to determine the superior methodology. Our statistical analysis assessed whether the correlation between feature $X$ and visual complexity differs significantly from that between feature $Y$ and complexity, allowing us to identify the more effective approach.

The permutation test first calculates Spearman's rank correlations ($r_s$) between complexity $C$ and both features $X$ and $Y$, computing the difference of their absolute values, denoted as $\Delta_{\text{obs}} = |r_s(C,X)| - |r_s(C,Y)|$. It then performs $n=1000$ permutations, each time randomly shuffling the complexity values $C^{(i)}$ while keeping feature values fixed, which breaks any real relationships between complexity and the features. For each permutation $i$, it calculates new correlations with the shuffled complexity values and records their permuted difference as $\Delta^{(i)}_{\text{perm}} = |r_s(C^{(i)},X)| - |r_s(C^{(i)},Y)|$. Statistical significance is determined by calculating a p-value:

$$p = \frac{\sum_{i=1}^{n} \mathbb{I}(|\Delta^{(i)}_{\text{perm}}| \geq |\Delta_{\text{obs}}|) + 1}{n + 1}$$

where $\mathbb{I}(\cdot)$ is the indicator function that equals 1 when the condition is true and 0 otherwise.A low $p$-value ($p < 0.05$) indicates the difference between correlations is statistically significant, and if significant, the feature with the higher absolute correlation is declared the \enquote{winner}.

We carried out a preliminary small-scale ablation study on weights and scales for MSG/MUC algorithms. Our baseline used scales $[1, 2, 4, 8]$ with weights $[0.4, 0.3, 0.2, 0.1]$ (\textit{baseline}). We compared this against three alternatives: (1) same scales with reversed weights $[0.1, 0.2, 0.3, 0.4]$ (\textit{alternative \#1}); (2) fewer scales $[1, 4]$ with weights $[0.6, 0.4]$ (\textit{alternative \#2}); (3) linearly incremental scales $[1, 2, 3, 4]$ with baseline weights (\textit{alternative \#3}) and (4) larger scales $[8, 16, 32, 64]$ with baseline weights (\textit{alternative \#4}). For MSG, we also evaluated using a larger $5 \times 5$ kernel size. For MSG, each comparison is carried out 16 times (number of datasets). For MUC, we carried out the comparison separately for each bit setting $\in \{3,4,5,6,7,8\}$ and reported the total number of win/loss/non-significant ratio ($16 \times 6 = 96$ runs).

\begin{table}
\centering{
\begin{tabular}{c|c|c}
\textbf{Feature} & \textbf{Comparison} & \textbf{Result (w/l/ns)} \\
\midrule
\multirow{5}{*}{MSG}
&\textit{baseline}  vs. \textit{alt. \#1}  & 7/4/5 \\
&\textit{baseline}  vs. \textit{alt. \#2}  & 4/3/9 \\
&\textit{baseline} vs.  \textit{alt. \#3} & 4/4/8 \\
&\textit{baseline} vs. \textit{alt. \#4}  & 10/1/5 \\
&\textit{baseline} vs. \textit{alt. kernel}  & 6/3/7 \\
\hline
\\[-8pt] 
\multirow{4}{*}{MUC} 
&\textit{baseline}  vs.  \textit{alt. \#1}  & 19/17/60 \\
&\textit{baseline} vs.  \textit{alt. \#2} & 22/10/64 \\
&\textit{baseline} vs.  \textit{alt. \#3} & 13/8/75 \\
&\textit{baseline} vs.  \textit{alt. \#4} & 51/9/36 \\
\hline
\end{tabular}}
\captionsetup{justification=centering}
\caption{Results of our preliminary ablation studies. Numbers indicate datasets where the first feature significantly outperformed (\textbf{w}in), underperformed (\textbf{l}oss), or showed no significant difference (\textbf{n}on-\textbf{s}ignificant) compared to the second feature ($p < 0.05$).}
\label{table:ws_ablation}
\end{table}

Based on our ablation study, we selected the \textit{baseline} weight and scale configuration for MSG and MUC algorithms to conduct subsequent experiments. While our ablation study guided the selection of a baseline configuration, we recognize that many other parameter combinations could be explored. A more heavily tuned setup might yield even better results, but we opted for a simpler baseline to focus on the broader methodological implications without spending excessive time on hyperparameter optimization.

We then extended our investigation by examining various algorithmic variations, including colored versus grayscale MSG, single-scale versus multi-scale implementations, and comparative analyses of different feature extraction techniques. These comparisons encompassed edge density and patch-based symmetry methods previously introduced by Davidson et al. \cite{KYLEDAVIDSON2023105319}. Table~\ref{table:ablation} highlights the effectiveness of colored MSG, multi-scale techniques, and MSG over the baseline alternatives.

\begin{table}
\centering{
\begin{tabular}{c|c}
\textbf{Comparison} & \textbf{Result (w/l/ns)} \\
\midrule
Colored MSG vs. Grayscale MSG & 6/2/8 \\
Multi-scale MSG vs. Single-scale MSG & 8/2/6 \\
Multi-scale MUC vs. Single-scale MUC & 5/1/10 \\
MSG vs. Edge Density & 5/2/9 \\
MSG vs. Patch-symmetry  & 7/1/8 \\
\hline
\end{tabular}}
\captionsetup{justification=centering}
\caption{Results of feature comparison ablation studies. Numbers indicate datasets where the first feature significantly outperformed (win), underperformed (loss), or showed no significant difference (non-significant) compared to the second feature ($p < 0.05$).}
\label{table:ablation}
\end{table}

\subsection{Conclusion}
\label{ap:conclusion}

In this work, we introduce novel visual features that significantly outperform those traditionally employed in visual complexity literature, enhancing the existing segmentation-based model while preserving its interpretability. Drawing inspiration from the familiarity bias, we explore a previously unexamined dimension of visual complexity: the element of surprise. We first developed the Surprising Visual Genome (SVG) dataset, created based on the well-studied Visual Genome collection, to systematically investigate semantic factors in complexity perception. Using this resource, we conducted human participant studies that revealed a consistent tendency to perceive unfamiliar, surprising, or bizarre images as visually more complex. We then used LLMs to automate the generation of surprise scores with explicit reasoning capabilities, creating a scalable approach to quantifying this subjective dimension of visual complexity without sacrificing interpretability. Our final model—which integrates low-level visual features, segmentation and class counts, and LLM-generated surprise scores—not only outperforms all previous handcrafted feature approaches but also achieves competitive results against the best-supervised model while maintaining interpretability.

\end{document}